\documentclass[conference]{IEEEtran}
\IEEEoverridecommandlockouts
\usepackage{cite}
\usepackage{amsmath,amssymb,amsfonts}
\usepackage{graphicx}
\usepackage{textcomp}
\usepackage{xcolor}
\def\BibTeX{{\rm B\kern-.05em{\sc i\kern-.025em b}\kern-.08em
    T\kern-.1667em\lower.7ex\hbox{E}\kern-.125emX}}

\usepackage{xspace}
\usepackage{makecell}
\usepackage{arydshln}
\usepackage{multirow}
\usepackage{amsmath}
\usepackage{relsize}
\usepackage{gensymb}
\usepackage{url}
\usepackage{adjustbox,booktabs,enumitem}
\usepackage{amsfonts,amssymb}
\usepackage{algorithm,algpseudocode}
\usepackage{subcaption}
\newcommand{\domain}[1]{\textit{#1}\xspace}
\newcommand{\dmbm}{\domain{biomedical}}
\newcommand{\dmcp}{\domain{computing}}
\newcommand{\dmms}{\domain{music}}
\newcommand{\dmfn}{\domain{finance}}
\newcommand{\dmfm}{\domain{film}}
\newcommand{\dmlw}{\domain{law}}

\newcommand{\dataset}[1]{\textsc{#1}\xspace}
\newcommand{\msmarco}{\dataset{msmarco}}
\newcommand{\ms}{\dataset{ms}}
\newcommand{\msbm}{\ms{}\dataset{-bm}}
\newcommand{\mscp}{\ms{}\dataset{-cp}}
\newcommand{\msfm}{\ms{}\dataset{-fm}}
\newcommand{\msfn}{\ms{}\dataset{-fn}}
\newcommand{\mslw}{\ms{}\dataset{-lw}}
\newcommand{\msms}{\ms{}\dataset{-ms}}
\newcommand{\bioasq}{\dataset{bioasq}}
\newcommand{\squad}{\dataset{squad}}
\newcommand{\method}[1]{\texttt{#1}\xspace}
\newcommand{\qanet}{\method{QANet}}
\newcommand{\bert}{\method{BERT}}
\newcommand{\bertl}{\method{BERT-Large}}
\newcommand{\bertb}{\method{BERT-Base}}
\newcommand{\deca}{\method{decaNLP}}
\newcommand{\scratch}{\method{scratch}}
\newcommand{\finetune}{\method{finetune}}
\newcommand{\ewc}{\method{+ewc}}
\newcommand{\ewcn}{\method{+ewcn}}

\newcommand{\cd}{\method{+cd}}
\newcommand{\ltwo}{\method{+l2}}
\newcommand{\all}{\method{+all}}
\newcommand{\gem}{\method{gem}}
\newcommand{\bb}[1]{\textbf{#1}}

\newcommand{\tabref}[2][]{Table#1~\ref{tab:#2}}
\newcommand{\figref}[2][]{Figure#1~\ref{fig:#2}}

\begin{document}

\title{Forget Me Not: Reducing Catastrophic Forgetting for Domain Adaptation in Reading Comprehension}

\author{\IEEEauthorblockN{1\textsuperscript{st} Ying Xu}
\IEEEauthorblockA{\textit{IBM Research Australia} \\
Melbourne, Australia \\
ying.jo.xu@au1.ibm.com}
\and
\IEEEauthorblockN{2\textsuperscript{nd} Xu Zhong}
\IEEEauthorblockA{\textit{IBM Research Australia} \\
Melbourne, Australia \\
peter.zhong@au1.ibm.com}
\and
\IEEEauthorblockN{3\textsuperscript{rd} Antonio Jose Jimeno Yepes}
\IEEEauthorblockA{\textit{IBM Research Australia} \\
Melbourne, Australia \\
antonio.jimeno@au1.ibm.com}
\and
\IEEEauthorblockN{4\textsuperscript{th} Jey Han Lau}
\IEEEauthorblockA{\textit{University of Melbourne} \\
Melbourne, Australia \\
jeyhan.lau@gmail.com}
}

\maketitle

\begin{abstract}
The creation of large-scale open domain reading comprehension data sets 
in recent years has enabled the development of end-to-end neural 
comprehension models with promising results. To use these models for 
domains with limited training data, one of the most effective approach 
is to first pre-train them on large out-of-domain \emph{source} data and 
then fine-tune them with the limited \emph{target} data.  The caveat of 
this is that after fine-tuning the comprehension models tend to perform 
poorly in the source domain, a phenomenon known as catastrophic 
forgetting.
In this paper, we explore methods that reduce catastrophic forgetting 
during fine-tuning without assuming access to data from the source 
domain.  We introduce new auxiliary penalty terms and observe the best 
performance when a combination of auxiliary penalty terms is used to 
regularise the fine-tuning process for adapting comprehension models. To 
test our methods, we develop and release 6 narrow domain data sets that 
can potentially be used as reading comprehension benchmarks. 
\end{abstract}


\section{Introduction}
Reading comprehension (RC) is the task of answering a question given a 
context passage. Related to Question-Answering (QA), RC is 
seen as a module in the full QA pipeline, where it assumes a related 
context passage has been extracted and the goal is to produce an answer 
based on the context. In recent years, the creation of large-scale open 
domain comprehension data sets 
\cite{Yang+:2015,nguyen+:2016,rajpurkar+:2016,Joshi+:2017,Rajpurkar+:2018,Kocisky+:2018} 
has spurred the development of a host of end-to-end neural comprehension 
systems with promising results.

In spite of these successes, it is difficult to train these modern 
comprehension systems on narrow domain data (e.g.\ biomedical), as these 
models often have a large number of parameters. A better approach is to 
transfer knowledge via fine-tuning, i.e.\ by first pre-training the 
model using data from a large \emph{source domain} and continously training 
it with examples from the small \emph{target domain}.  It is an 
effective strategy, although a fine-tuned model often performs poorly 
when it is re-applied to the source domain, a phenomenon known as 
catastrophic forgetting 
\cite{french1999catastrophic,wiese+:2017,kirkpatrick+:2017,riemer2018learning}.  
This is generally not an issue if the goal is to optimise purely for the 
target domain, but in real-word applications where model robustness is 
an important quality, over-optimising for a development set often leads 
to unexpected poor performance when applied to test cases in the wild.

Methods for reducing catastrophic forgetting are categorised into three types
based on three assumptions, i.e. with \textit{full}, \textit{partial} or \textit{no} access to the source
domain data. When \textit{full} access is available, the 
most straightforward way to reduce catastrophic forgetting is to perform 
multitask learning, where the model is learned to perform well for both source
and target domains \cite{Daume:2007,kim2016frustratingly}. 
For \textit{partial} access assumption, the gradient episodic memory 
(GEM) \cite{lopez2017gradient} was proposed to store a piece of data from 
the source domain, which is used to regularise the fine tuning process. 
However, in real-world applications, data became inaccessible for a number of 
reasons, such as an expired data sharing agreement, physical damage to data 
storage, accidental deletion of data, and the introduction of new data privacy
laws. Here we focus on exploring methods to reduce catastrophic forgetting
assuming \textit{no} access to the source domain data. 


In this paper, we explore strategies to reduce forgetting for comprehension 
systems during domain adaptation. Our goal is to preserve the source domain's 
performance as much as possible, while keeping target domain's 
performance optimal and assuming no access to the source data. We 
experiment with a number of auxiliary penalty terms to regularise the 
fine-tuning process for three modern RC models: \qanet
\cite{yu+:2018}, \deca \cite{mccann+:2018} and \bert \cite{devlin2018bert}. 
We observe that combining different auxiliary penalty terms results
in the best performance, outperforming benchmark methods that require 
source data.
Technically speaking, the application of the methods we propose are not limited to domain 
transfer for reading comprehension. We also show that the methodology 
can be used for transferring to entirely different natural language processing tasks. 
With that said, we focus on comprehension here because it is a practical problem 
in real world applications, where the target domain often has a small 
number of QA pairs and over-fitting occurs easily when we fine-tune 
based on a small development set. In this scenario, it is as important 
to develop a robust model as achieving optimal development performance.

To demonstrate the applicability of our approach, we apply topic 
modelling to \msmarco \cite{nguyen+:2016} --- a comprehension data set 
based on internet search queries --- and collect examples that belong to 
a number of salient topics, producing 6 small to medium sized RC data 
sets for the following domains: \dmbm, \dmcp, \dmfm, \dmfn, \dmlw and 
\dmms.
We focus on extractive RC, where the answer is a continuous sub-span in 
the context passage.\footnote{Although RC with free-form answers is 
arguably a more challenging and interesting task, evaluation is 
generally more difficult \cite{Kwiatkowski+:2019}.}
Scripts to generate the data sets are available at: 
\url{https://github.com/ibm-aur-nlp/domain-specific-QA}.


Summarising our contributions: (1) we experiment with a number of 
penalty terms to regularise the fine-tuning process for adapting 
comprehension systems, and found that a combination of them produces the 
most robust model that both preserves performance in the source domain 
and achieves optimal performance in the target domain; and (2) we 
develop and release six narrow-domain extractive comprehension data 
sets, facilitating research on domain adaptation for reading 
comprehension.

\section{Related Work}

Most large comprehension data sets are open-domain because non-experts 
can be readily recruited via crowdsourcing platforms to collect annotations. 
Development of domain-specific RC data sets, on the other hand, is 
costly due to the need of subject matter experts and as such the size of 
these data sets is typically limited. Examples include 
\bioasq~\cite{tsatsaronis2015overview} in the biomedical domain, which 
has less than 3k QA pairs --- orders of magnitude smaller compared to 
most large-scale open-domain data sets 
\cite{nguyen+:2016,rajpurkar+:2016,Joshi+:2017,Kocisky+:2018}.

Wiese et al. \cite{wiese+:2017} explore supervised domain adaptation for reading 
comprehension, by pre-training their model first on large open-domain 
comprehension data and fine-tuning it further on biomedical data. This 
approach improves the biomedical domain's performance substantially 
compared to training the model from scratch.  At the same time, its 
performance on source domain decreases dramatically due to catastrophic 
forgetting 
\cite{french1999catastrophic,mccloskey1989catastrophic,riemer2017representation}.

This issue of catastrophic forgetting is less of a problem when data 
from multiple domains or tasks are present during training. For example 
in \cite{mccann+:2018}, their model \deca is trained on 10 tasks 
simultaneously --- all cast as a QA problem --- and forgetting is 
minimal.  For multi-domain adaptation,  Daumé III, H. \cite{Daume:2007} and 
Kim et al. \cite{kim2016frustratingly} propose using a K+1 model to capture 
domain-general pattern that is shared by K domains, resulting in a more 
robust model.  Using multi-task learning to tackle catastrophic 
forgetting is effective and generates robust models. The drawback,   
however, is that when training for each new domain/task, data from the 
previous domains/tasks has to be available. 


Several studies present methods to reduce forgetting with limited
or no access to previous data~\cite{riemer2017scalable,lopez2017gradient,kirkpatrick+:2017,serra2018overcoming,riemer2018learning}.  
Inspired by 
synaptic consolidation, Kirkpatrick et al. \cite{kirkpatrick+:2017} propose to 
selectively penalise parameter change during fine-tuning. Significant 
updates to parameters which are deemed important to the source task 
incur a large penalty.  
Lopez-Paz et al. \cite{lopez2017gradient} introduce a gradient episodic memory (\gem) to 
allow beneficial transfer of knowledge from previous tasks. More specifically,
a subset of data from previous tasks are stored in an episodic memory, 
against which reference gradient vectors are calculated and the angles 
with the gradient vectors for the current task is constrained to be between
$-90\degree$ and $90\degree$.  Riemer et al. \cite{riemer2018learning} suggest combining \gem
with optimisation based meta-learning to overcome forgetting. Among 
these three methods, only that of \cite{kirkpatrick+:2017} assumes zero 
access to previous data. In comparison, the latter two rely on access to 
a memory storing data from previous tasks, which is not always feasible 
in real-world applications (, as is mentioned in the previous section).






\begin{table}
  \begin{center}
  \begin{adjustbox}{max width=\linewidth}
  \begin{tabular}{ccrrrrr}
    \toprule
    \multirow{2}{*}{\textbf{Partition}} & 
\multirow{2}{*}{\textbf{Domain}} & \multirow{2}{*}{\textbf{\#Examples}} 
& \multirow{2}{*}{\textbf{\#Unique Q}} & \textbf{Mean C} & \textbf{Mean 
Q} & \textbf{Mean A} \\
&&&& \textbf{Length}  & \textbf{Length} & \textbf{Length} \\

    \midrule
    \multirow{ 7}{*}{Train} & \msbm & 22,134 & 21,902 & 70.9 & 6.4 & 13.7 \\
    & \mscp & 3,021 & 3,011 & 67.2 & 5.5 & 18.9 \\
    & \msfm & 3,522 & 3,481 & 65.8 & 6.4 & 6.5 \\
    & \msfn & 6,790 & 6,720 & 71.9 & 6.4 & 14.0 \\
    & \mslw & 3,105 & 3,078 & 64.7 & 6.2 & 18.5 \\
    & \msms & 2,517 & 2,480 & 68.6 & 6.4 & 6.6 \\
    & \bioasq & 3,083 & 387 & 35.4 & 11.0 & 2.4 \\
    \midrule
    \multirow{ 7}{*}{Dev} & \msbm & 4,743 & 4,730 & 71.2 & 6.4 & 13.7 \\
    & \mscp & 647 & 646 & 65.4 & 5.3 & 19.6 \\
    & \msfm & 755 & 751 & 65.9 & 6.6 & 5.9 \\
    & \msfn & 1,455 & 1,453 & 71.6 & 6.5 & 14.4 \\
    & \mslw & 665 & 664 & 65.8 & 6.2 & 20.0 \\
    & \msms & 539 & 536 & 69.2 & 6.4 & 6.1 \\
    & \bioasq & 674 & 83 & 39.7 & 11.1 & 2.4 \\
    \midrule
    \multirow{ 7}{*}{Test} & \msbm & 4,743 & 4,728 & 70.5 & 6.4 & 13.5 \\
    & \mscp & 648 & 645 & 66.6 & 5.6 & 18.3 \\
    & \msfm & 755 & 755 & 66.7 & 6.3 & 6.2 \\
    & \msfn & 1,455 & 1,452 & 70.8 & 6.5 & 13.6 \\
    & \mslw & 666 & 663 & 65.1 & 6.2 & 18.9 \\
    & \msms & 540 & 540 & 67.4 & 6.6 & 7.0 \\
    & \bioasq & 631 & 84 & 34.9 & 13.2 & 2.9 \\
    \bottomrule
  \end{tabular}
  \end{adjustbox}
  \end{center}
  \caption{Statistics of our seven target domain data sets (Q: Question; C: Context; and A: Answer).}
  \label{tab:stat}
\end{table}

\section{Data Set}
\label{sec:dataset}

We use \squad v1.1 \cite{rajpurkar+:2016} as the source domain data for 
pre-training the comprehension model.  It contains over 100K extractive (context, 
  question, answer) triples with only answerable questions.

To create the {target domain} data, we leverage \msmarco 
\cite{nguyen+:2016}, a large RC data set where questions are sampled 
from Bing\textsuperscript{\texttrademark} search
queries and answers are manually generated by users based on passages in web
documents. We apply LDA topic model \cite{Blei:2003} to passages in \msmarco and
learn 100 topics.\footnote{When collecting the passages, we include only those
being selected as useful for answering a query (i.e.\ \texttt{is\_selected} $= 1$).  We tokenise the
passages with Stanford CoreNLP \cite{Manning+:2014} and use MALLET
\cite{McCallum:2002} for topic modelling.} Given the topics, we label 
 them and select 6 salient domains: \dmbm (\msbm), \dmcp 
(\mscp), \dmfm (\msfm), \dmfn (\msfn), \dmlw (\mslw) and \dmms (\msms). A 
QA pair is categorised into one of these domains if its passage's 
top-topic belongs to them. We create multiple (context, question, 
answer) training examples if a QA pair has multiple 
contexts,\footnote{We only consider context passages that are marked as 
being useful by annotators in the original data (i.e.\ 
\texttt{is\_selected} $= 1$).}  and filter them to keep only extractive 
examples.\footnote{A (context, question, answer) triple is defined to be 
extractive if the
answer has a case-insensitive match to the context.} 

In addition to the \msmarco data sets, we also experiment with a real 
biomedical comprehension data set: \bioasq~\cite{tsatsaronis2012bioasq}.  
Each question in \bioasq is associated with a set
of snippets as context, and the snippets are single sentences 
extracted from a scientific publication's abstract/title in PubMed 
Central\textsuperscript{\texttrademark}. There are four types of 
questions: factoid, list, yes/no, and summary. As our focus is on 
extractive RC, we use only the extractive factoid questions from 
\bioasq. As before, we create multiple training examples for QA pairs 
with multiple contexts.

For each target domain, we split the examples into 70\%/15\%/15\% 
training/development/test partitions.\footnote{Partitioning is done at 
the question level to ensure the same question does not appear in more 
than one partition.} We present some statistics for the data sets in 
\tabref{stat}. 

\section{Methodology}

We first pre-train a general domain RC model on \squad, our {source 
domain}.  Given the pre-trained model, we then perform fine-tuning 
(\finetune) on the
\msmarco and \bioasq data sets: 7 {target domains} in total.  By 
fine-tuning we mean taking the pre-trained model parameters as initial 
parameters and update them  accordingly based on data from the new 
domain.  To reduce forgetting on
the source domain (\squad), we experiment with incorporating auxiliary
penalty terms (e.g.\ L2 between new and old parameters) to the standard 
cross entropy loss to regularise the fine-tuning
process.

We explore 3 modern RC models in our experiments: \qanet
\cite{yu+:2018}; \deca \cite{mccann+:2018}; and \bert \cite{devlin2018bert}. 
\qanet is a Transformer-based \cite{Vaswani+:2017} comprehension model, 
where the encoder consists of stacked convolution and self-attention layers. 
The objective of the model is to predict the position of the starting and
ending indices of the answer words in the context.  \deca is a recurrent 
network-based comprehension model trained on ten NLP tasks 
simultaneously, all casted as a question-answer problem. Much of \deca's 
flexibility is due to its pointer-generator network, which allows it to 
generate words by extracting them from the question or context passages, 
or by drawing
them from a vocabulary.  \bert is a deep bi-directional encoder model 
based on Transformers. It is pre-trained on a large corpus in an 
unsupervised fashion using a masked language model and next-sentence 
prediction objective. To apply \bert to a specific task, the standard 
practice is to add additional output layers on top of the pre-trained 
\bert and fine-tune the whole model for the task. In our case for RC, 
2 output layers are added: one for predicting the start index and 
  another the end index.  Devlin et al. \cite{devlin2018bert} demonstrates that this 
transfer learning strategy produces state-of-the-art performance on a 
range of NLP tasks.  For RC specifically, \bert (\bertl) achieved an F1 
score of 
93.2 on \squad, outperforming human performance by 2 points.

Note that \bert and \qanet RC models are extractive models (goal is to 
predict 2 indices), while \deca is a generative model (goal is to 
generate the correct word sequence).  Also, unlike \qanet and \deca, 
\bert is not designed specifically for RC. It represents a growing trend 
in the literature where large models are pre-trained on big corpora and 
further adapted to downstream tasks.  

To reduce the forgetting of source domain knowledge,  we introduce 
auxiliary penalty terms to regularise the fine-tuning process.  We 
favour this approach as it does not require storing data samples from 
the source domain.  In general, there are two types of penalty: 
selective and non-selective.  The former penalises the model when 
certain parameters diverge significantly from the source  model, while 
the latter uses a pre-defined distance function to measure the change of 
all parameters.

For selective penalty, we use elastic weight consolidation (EWC: 
\cite{kirkpatrick+:2017}), which weighs the importance of a parameter 
based on its gradient when training the source model. For non-selective 
penalty, we explore L2 \cite{wiese+:2017} and cosine distance.  We 
detail the methods below.


Given a source and target domain, we pre-train the model first on the 
 source domain and fine-tune it further on the target domain. We denote 
the optimised parameters of the source model as ${\theta^*}$ and that of 
the target model as ${\theta}$.  For vanilla fine-tuning (\finetune), 
the loss function is:
\begin{equation*}
    \mathcal{L}_{ft}=\mathcal{L}_{ce}
\end{equation*}
where $\mathcal{L}_{ce}$ is the cross-entropy loss.

For non-selective penalty, we measure the change of parameters based on 
a distance function (treating all parameters as equally important), and add 
it as a loss term in addition to the 
cross-entropy loss. One distance function we test is the L2 distance:
\begin{equation*}
    \mathcal{L}_{+l2} = \mathcal{L}_{ce} + \lambda_{l2} 
\text{L2}(\theta, \theta^*)
\end{equation*}
where $\lambda_{l2}$ is a scaling hyper-parameter to weigh the 
contribution of the penalty. Henceforth all scaling hyper-parameters are 
denoted using $\lambda$.

We also experiment with cosine distance, based on the idea that we want 
to encourage the parameters to be in the same direction after 
fine-tuning.  In this case, we group parameters by the variables they 
are defined in, and measure the cosine distance between variables:
\begin{equation*}
    \mathcal{L}_{+cd} = \mathcal{L}_{ce} + \lambda_{cd} \frac{1}{|V|} 
\sum_v \text{CD}(\theta_{v}, \theta^*_{v})
\end{equation*}
where $\theta_v$ denotes the vector of parameters belonging to variable 
$v$.

For selective penalty, EWC uses the Fisher matrix $F$ to measure the 
importance of parameter $i$ in the source domain. Unlike non-selective 
penalty where all parameters are considered equally important, EWC 
provides a mechanism to weigh the update of individual parameters:
\begin{align*}
    \mathcal{L}_{+ewc} &= \mathcal{L}_{ce} + \lambda_{ewc}
\sum_{i}(F_{i} \cdot (\theta_i - \theta^*_{i})) \\
{F} &= E[(\frac{\partial  \mathcal{L}_{ce} (f_{\theta^*}, (x, y))}{\partial \theta^*})^2|\theta^*]
\end{align*}
where $\frac{\partial  \mathcal{L}_{ce} (f_{\theta^*}, (x, y))}{\partial \theta^*}$ is the gradient 
of parameter update in the source domain, with $f_{\theta^*}$ representing the
model and $x$/$y$ the data/label from the source domain.

 \begin{figure}[t]
    \centering
    \begin{subfigure}[b]{0.23\textwidth}
        \includegraphics[width=1.0\linewidth]{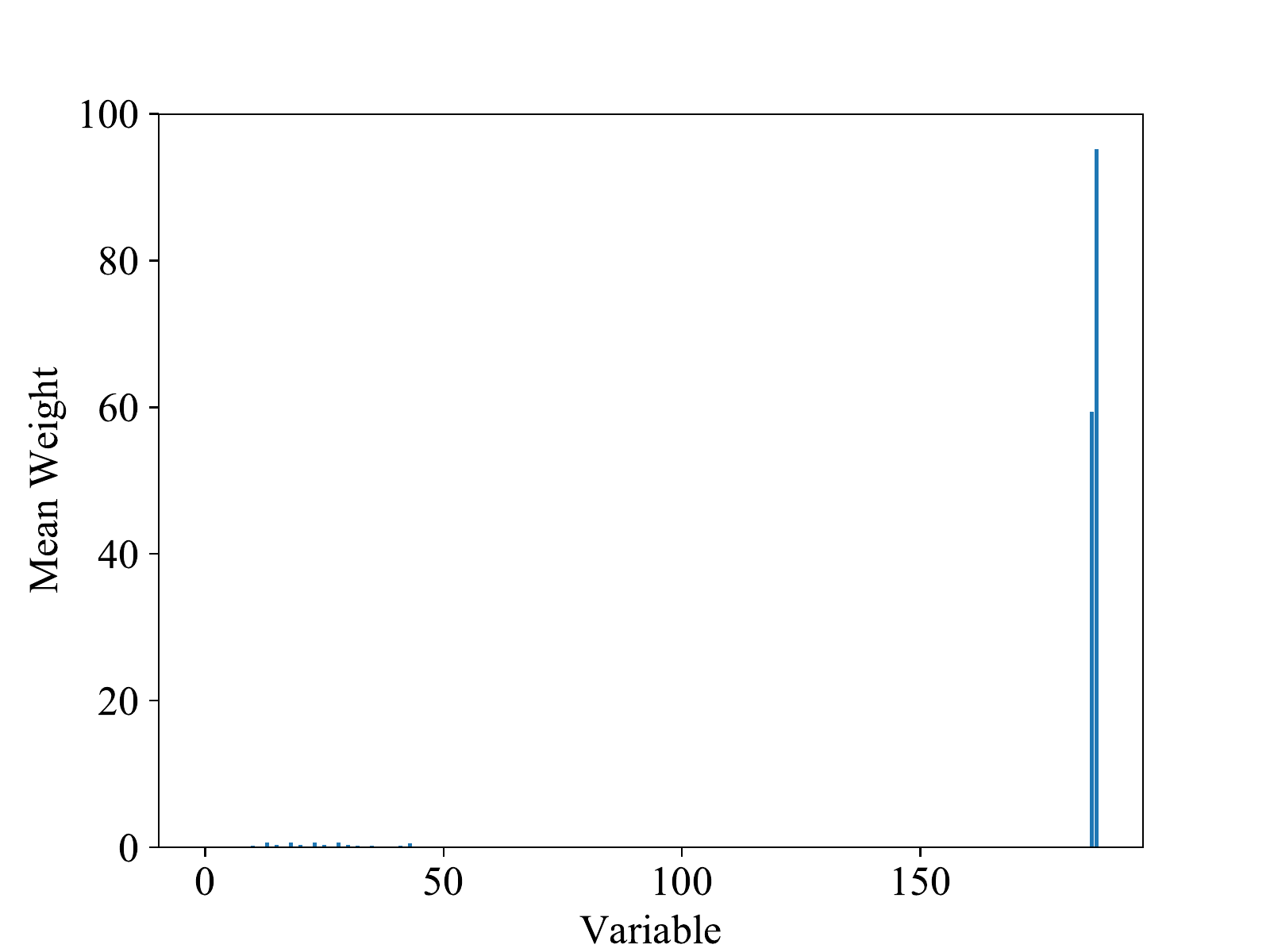}
        \caption{EWC}
        \label{fig:fisher-matrix}
    \end{subfigure}
    \begin{subfigure}[b]{0.23\textwidth}
        \includegraphics[width=1.0\linewidth]{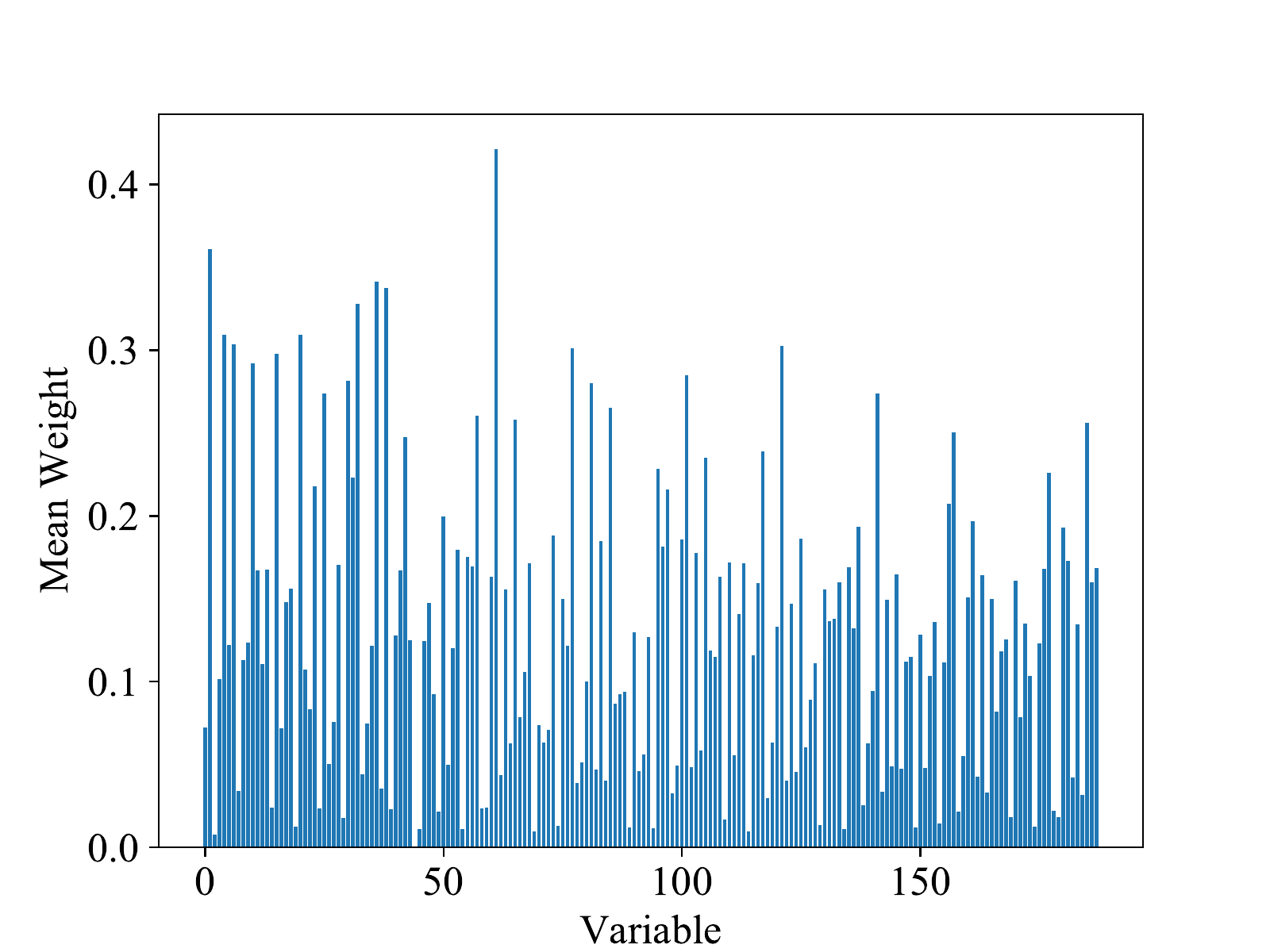}
        \caption{Normalised EWC}
	\label{fig:fisher-matrix-norm}
    \end{subfigure}
    \caption{Mean Fisher Matrix values for variables in \qanet on 
\squad.}
\end{figure}

%

In preliminary experiments, we notice that EWC tends to assign most of 
the weights to a small subset of parameters. We present 
\figref{fisher-matrix}, a plot of mean Fisher values for all variables 
\footnote{By variable we intend a parameter matrix in the attention or 
feed-forward layers, e.g. the kernel matrix of the 6th layer defined in a \bert 
model. }
in \qanet after it was trained on \squad, the source domain. We see that  
only the last two variables have some significant weights (and a tiny 
amount for the rest of the variables). 
Due to the considerably different contribution of variables in different layers
, using the Fisher matrix above directly may leads to ignoring the difference 
in the contribution of different parameters within the same variable. 
We therefore propose a new variation 
of EWC, \textit{normalised EWC}, by normalising the 
weights within each variable via min-max normalisation, which brings up 
the weights for parameters in other variables (\figref{fisher-matrix-norm}):
\begin{align*}
    {F}^{*}_{i} &= 
    \frac{{F}_{i}-min(\{F\}^{v_i})}{max(\{F\}^{v_i})-min(\{F\}^{v_i})} 
\\
    \mathcal{L}_{+ewcn} &= \mathcal{L}_{ce} + \lambda_{ewcn}
\sum_{i}(F^{*}_{i} \cdot (\theta_i - \theta^*_{i}))
\end{align*}
where $\{F\}^{v_i}$ denotes the set of parameters for variable $v$ where 
parameter $i$ belongs.

Among the four auxiliary penalty terms, L2 and EWC are proposed in 
previous work while cosine distance and normalised EWC are novel penalty 
terms. Observing that EWC and normalised EWC are essentially 
weighted $l1$ distances, L2 is based on $l2$ distance 
and cosine distance focuses on the vector angle between 
variables (and ignores the magnitude), we propose combining them 
altogether as these different distance metrics may complement
each other in regularising the fine-tuning process:

\begin{align*}
    \mathcal{L}_{+all} = \mathcal{L}_{ce} &+ \lambda_{l2} 
\text{L2}(\theta, \theta^*) \\
    &+ \lambda_{cd} \frac{1}{|V|} \sum_v \text{CD}(\theta_{v}, 
    \theta^*_{v}) \\
    &+ \lambda_{ewcn}
\sum_{i}(F^{*}_{i} \cdot (\theta_i - \theta^*_{i}))
\end{align*}

Since EWC and normalised EWC are both weighted $l1$ distances
and our results demonstrate that our proposed normalised EWC perform 
better than the vanilla EWC, we only incorporate normalised EWC in the 
combined loss $\mathcal{L}_{+all}$. 


\begin{table*}[t!]
\begin{center}
\begin{adjustbox}{max width=0.7\linewidth}
\begin{tabular}{ccccccccccc}
\toprule
\textbf{Model} & \textbf{Partition} & \textbf{Domain} & 
\textbf{\scratch} &
\textbf{\finetune} & \textbf{\ewc} & \textbf{\ewcn} & \textbf{\cd} &
\textbf{\ltwo} & \textbf{\all} & \textbf{\gem} \\
\midrule

\multirow{16}{*}{\textbf{\qanet}} & \multirow{8}{*}{\squad (dev)}
& \msbm & --- & 62.92 & 63.35 & 63.93 & 63.49 & 64.93 & \bb{65.54} &63.22 \\
&& \mscp & --- & 39.13 & 41.62 & 43.43 & 41.19 & 41.61 & \bb{51.84}& 43.49 \\
&& \msfm & --- & 56.32 & 58.23 & 58.46 & 57.01 & 58.48 & \bb{60.79} & 57.53 \\
&& \msfn & --- & 65.08 & 65.45 & 67.03 & 65.36 & 66.27 & \bb{68.14} & 66.53 \\
&& \mslw & --- & 68.29 & 68.64 & 68.63 & 68.75 & 68.38 & \bb{69.39}& 69.04 \\
&& \msms & --- & 69.60 & 69.96 & 70.11 & 69.72 & 69.74 & \bb{71.13}& 70.63 \\
&& \bioasq & --- & 59.85 & 62.87 & 63.57 & 62.83 & 62.50 & \bb{66.11}& 62.52 \\
\cdashline{3-11}
&& Avg. & --- & 60.17 & 61.45 & 62.17 & 61.19 & 61.70 & \bb{64.71} & 61.85 \\

\cline{2-11}

& \multirow{8}{*}{Target (test)}
& \msbm & 62.75 & \underline{68.45} & 67.96 & 67.85 & 67.80 & 68.05 & 67.33 & 68.31 \\
&& \mscp & 60.67 & 68.86 & 69.26 & 69.86 & 70.27 & 69.42 & \underline{70.42}& 69.17 \\
&& \msfm & 59.57 & 73.84 & 72.70 & \underline{74.13} & 73.94 & 73.50 & 73.47 & 72.00 \\
&& \msfn & 63.62 & \underline{70.96} & 70.70 & 70.60 & 70.49 & 70.15 & 70.27 & 69.18 \\
&& \mslw & 61.66 & 71.29 & 71.27 & 71.39 & 71.25 & 71.28 & 71.41& \underline{71.49} \\
&& \msms & 58.36 & 69.58 & 69.94 & 69.89 & 69.62 & 69.92 & 70.67 & \underline{71.47} \\
&& \bioasq & 29.83 & 65.81 & 67.17 & 65.93 & \underline{67.26} & 65.57 & 66.82& 66.42 \\
\cdashline{3-11}
&& Avg. & 56.64 & 69.83 & 69.86 & 69.95 & \underline{70.09} & 69.70 & 70.06 & 69.72 \\

\midrule

\multirow{16}{*}{\textbf{\deca}} & \multirow{8}{*}{\squad (dev)}
& \msbm & --- & 62.99 & 63.00 & 63.26 & 63.27 & 62.43 & 63.82 & \bb{64.95} \\
&& \mscp & --- & 56.48 & 58.19 & 59.44 & 61.96 & 60.73 & 62.61 & \bb{63.37} \\
&& \msfm & --- & 58.69 & 59.21 & 59.18 & 62.66 & 58.32 & \bb{64.04 } & 63.36 \\
&& \msfn & --- & 58.21 & 61.63 & 63.43 & 59.25 & 58.80 & \bb{66.55} & 62.47 \\
&& \mslw & --- & 57.86 & 58.14 & 59.73 & 58.17 & 56.89 & 60.75 & \bb{61.76} \\
&& \msms & --- & 59.75 & \bb{64.92} & 62.01 & 62.00 & 60.06 & 63.62 & 63.89 \\
&& \bioasq & --- & 67.42 & 67.19 & 67.21 & 67.44	& 67.46 & 67.49 & \bb{68.94} \\
\cdashline{3-11}
&& Avg. & --- & 60.20 & 61.75 & 62.04 & 62.11 & 60.67 & \bb{64.13} & 64.11 \\

\cline{2-11}

& \multirow{8}{*}{Target (test)}
& \msbm & 62.01 & 66.90 & 67.39 & 67.52 & \underline{67.61} & 67.19 & 67.41 &  67.02 \\
&& \mscp & 63.7 & 66.67 & 67.11 & \underline{68.15} & 66.37 & 67.82 & 67.55 &67.90 \\
&& \msfm  & 63.28 & 70.45 & 70.47 & \underline{70.83} & 69.08 & 70.36 & 68.04 & 69.73 \\
&& \msfn  & 64.41 & 64.59 & 64.57 & 64.35 & 64.32 & 64.87 & 64.32 & \underline{64.88} \\
&& \mslw  & 66.36 & 73.43 & 73.28 & 73.34 & 73.42 & \underline{74.13} & 73.04 & 72.89 \\
&& \msms  & 64.65 & 68.67 & 67.12 & 67.93 & 67.34 & \underline{69.40} & 66.51 & 68.28 \\
&& \bioasq  & 43.25 & 63.80 & 63.89 & 63.89 & 63.96 & 63.96 & 64.70 & \underline{66.36} \\
\cdashline{3-11}
&& Avg. & 61.09 & 67.79 & 67.69 & 68.00 & 67.44 & \underline{68.25} & 67.37 & 68.15 \\

\midrule

\multirow{16}{*}{\textbf{\bert}} & \multirow{8}{*}{\squad (dev)}
& \msbm & --- & 72.55	& 74.24	& 76.51	& 72.36	& 74.14	& \bb{77.32} & 74.14 \\
&& \mscp & --- & 68.41	& 69.63	& 75.65	& 76.92	& 75.98	& \bb{77.86} & 73.37 \\
&& \msfm & --- & 73.82	& 75.175	& 79.75	& 75.28	& 74.71	& \bb{81.42} & 76.89 \\
&& \msfn & --- & 72.59	& 74.27	& 75.52	& 73.22	& 74.84	& \bb{78.18} & 76.16\\
&& \mslw & --- & 71.93	& 81.11	& 81.05	& 78.77	& 77.97	& \bb{83.11} & 75.90 \\
&& \msms & --- & 72.59	& 78.06	& \bb{83.56}	& 75.67	& 74.29	& 83.54 & 76.99\\
&& \bioasq & --- & 75.04	& 85.28	& 85.62	& 85.76	& 84.23	& \bb{86.88} & 75.89 \\
\cdashline{3-11}
&& Avg. & --- & 72.42 & 76.82 & 79.67 & 76.85 & 76.59 & \bb{81.19} & 75.62 \\

\cline{2-11}

& \multirow{8}{*}{Target (test)}
& \msbm & 66.83 & \underline{68.30}	& 68.20	&68.00	&68.04	&68.24	&67.87 & 68.02 \\
&& \mscp & 65.99 & 70.57	& 71.21	& \underline{71.41}	& 69.33	& 69.57	& 69.49 & 70.40 \\
&& \msfm & 72.59 & 74.73	& 74.75	& 74.36	& 73.73	& 74.85	& \underline{75.78} & 74.63 \\
&& \msfn  & 66.70 & 69.13 & 70.42	& \underline{70.60}	& 69.07	& 70.05	& 69.15 & 69.54 \\
&& \mslw & 67.38 & 69.99	& 70.73	& \underline{71.59}	& 70.57	& 70.91	& 68.59 & 68.87\\
&& \msms & 70.45 & \underline{73.56} & 73.19	& 73.07	& 72.97	& 73.43	& 72.50 & 72.73\\
&& \bioasq & 54.09 & 71.62 	& 75.84	& 78.50	& \underline{79.47}	& 78.86	& 76.93 & 68.87 \\
\cdashline{3-11}
&& Avg. & 66.29 & 71.13 & 72.05 & \underline{72.50} & 71.88 & 72.27 & 71.47 & 70.44 \\

\bottomrule
\end{tabular}
\end{adjustbox}
\end{center}
\caption{RC results over all domains. Pre-trained \qanet/\deca/\bert 
performance on \squad (dev) $=$ 80.47/75.50/87.62. Boldface indicates optimal 
performance for \squad (dev) and Underline indicates best performance for target 
domains (test).}
\label{tab:rc-full}
\end{table*}

\begin{figure*}
    \centering
    \begin{subfigure}[b]{0.27\textwidth}
        \includegraphics[width=\textwidth]{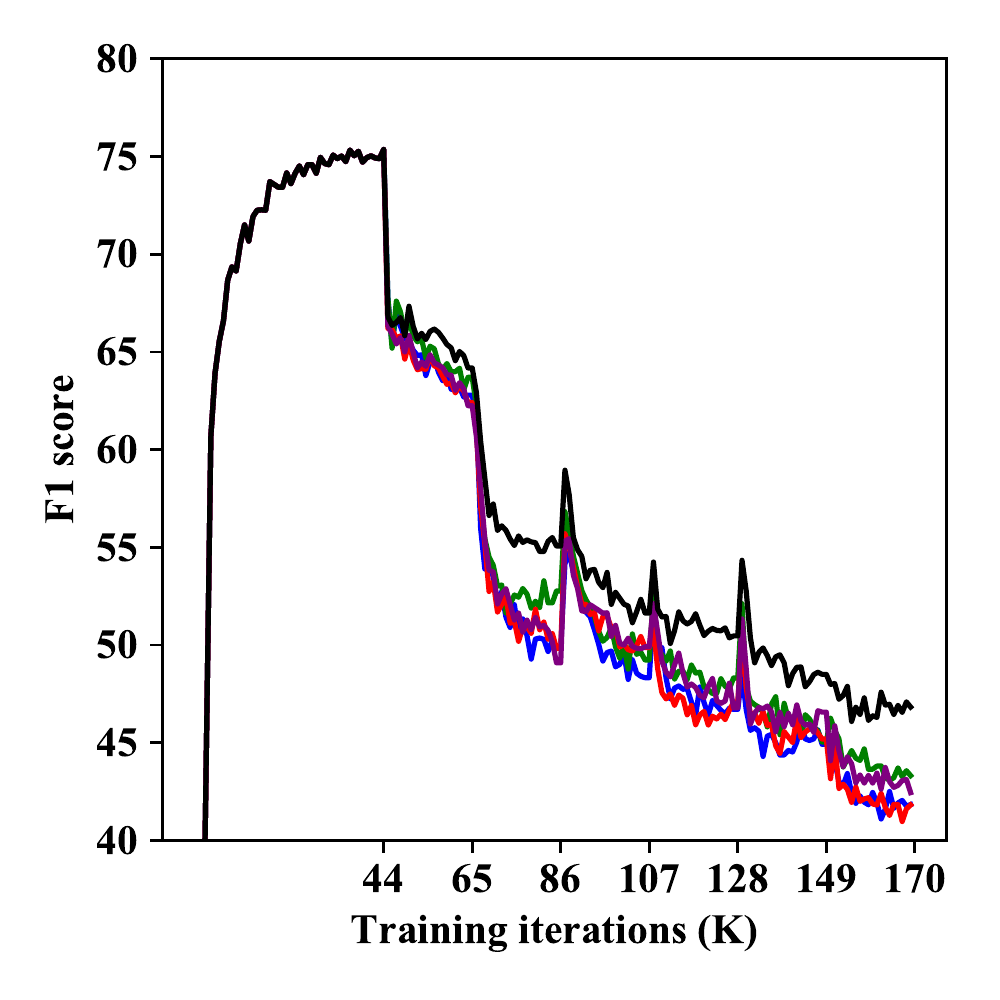}
        \caption{\squad}
        \label{fig:cont_squad}
    \end{subfigure}
    ~ 
    \begin{subfigure}[b]{0.27\textwidth}
        \includegraphics[width=\textwidth]{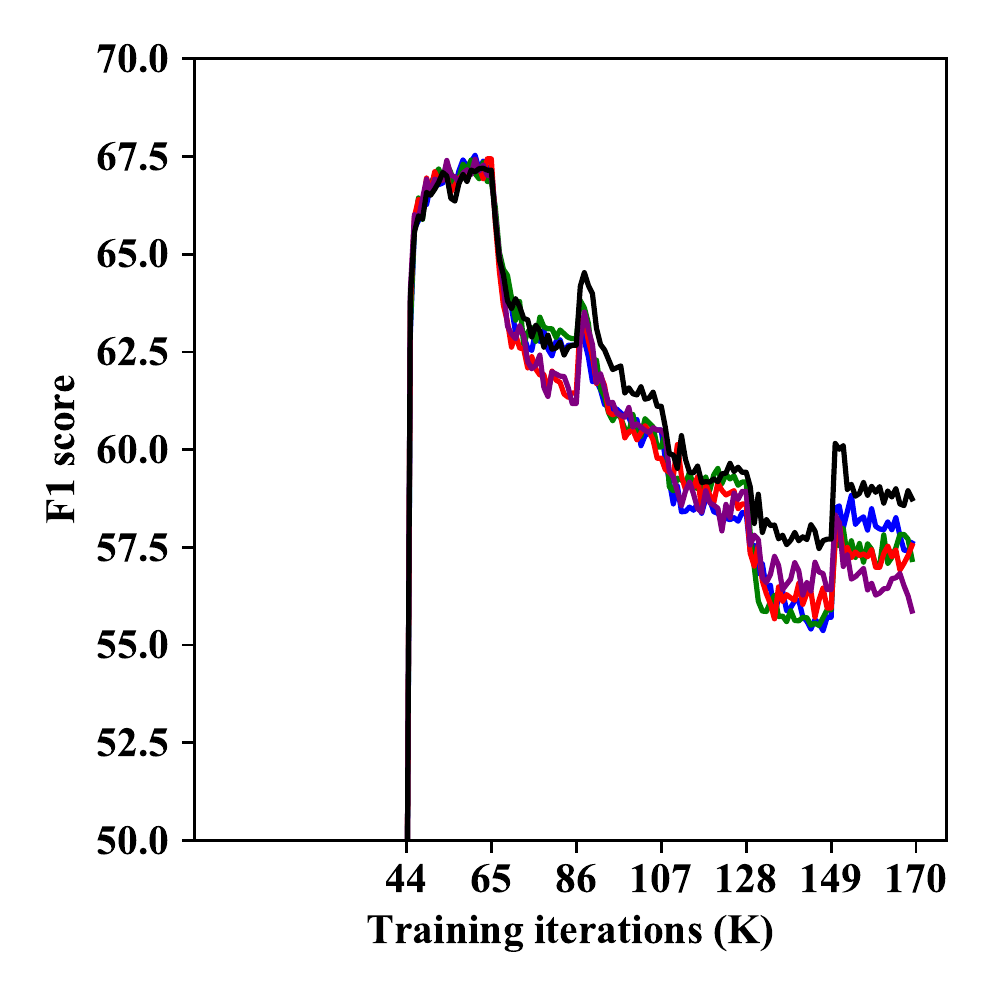}
        \caption{\msbm}
        \label{fig:cont_bio}
    \end{subfigure}
    ~ 
    \begin{subfigure}[b]{0.27\textwidth}
        \includegraphics[width=\textwidth]{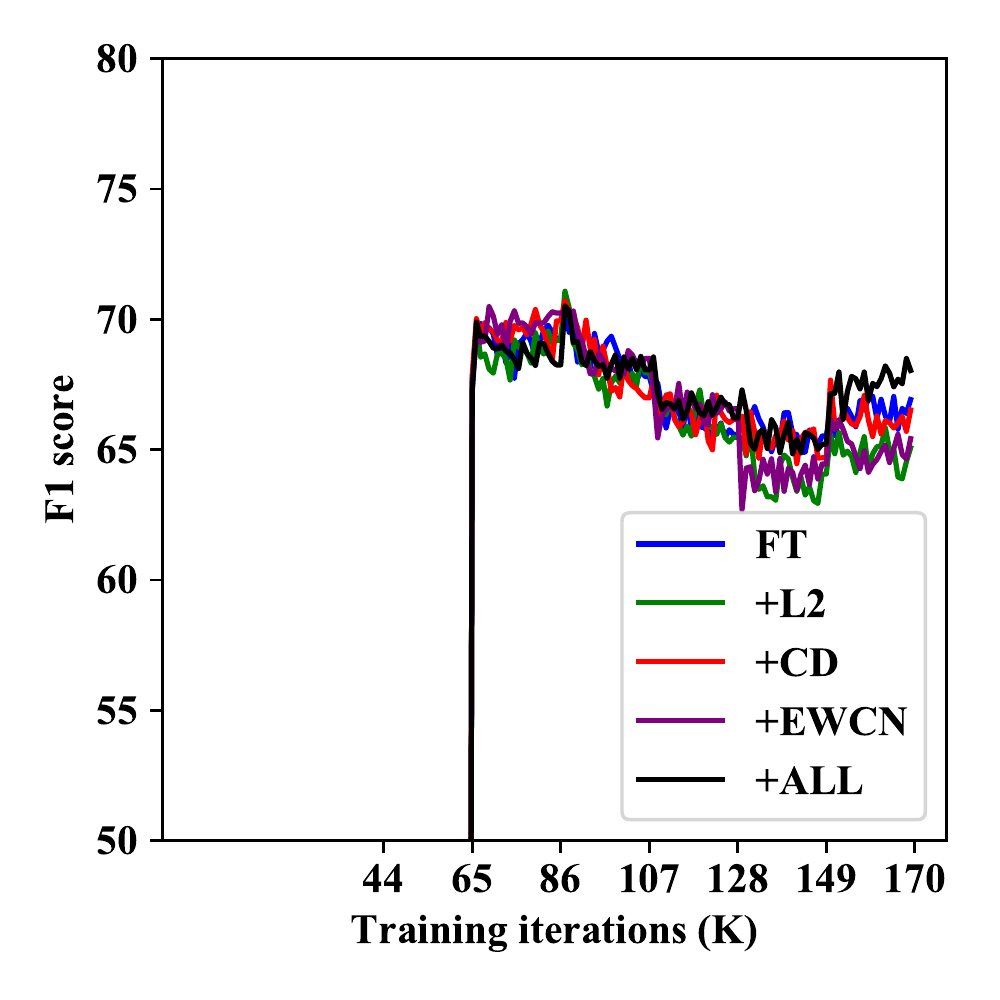}
        \caption{\mscp}
        \label{fig:cont_comp}
    \end{subfigure}
    \caption{\deca's F1 performance during continuous learning.}
    \label{fig:continuous}
\end{figure*}

\section{Experiments}

We test 3 comprehension models: \qanet, \deca and \bert. To pre-process 
the data, we use the the models' original tokenisation 
methods.\footnote{We use spaCy (\url{https://spacy.io/}), 
revtok (\url{https://github.com/jekbradbury/revtok}), and WordPiece for 
\qanet, \deca and \bert, respectively.} For \bert, we use the smaller 
pre-trained model with 110M parameters (\bertb).

\subsection{Fine-Tuning with Auxiliary Penalty}

We first pre-train \qanet and \deca on \squad, tuning their 
hyper-parameters based on its development partition.\footnote{We tune 
for dropout, batch size, learning rate and number of training 
iterations, keeping other hyper-parameters in their default 
configuration.} For \bert, we fine-tune the released pre-trained model 
on \squad by adding 
2 additional output layers to predict the start/end indices (we made no 
  changes to the hyper-parameters).  We initialise word vectors of 
\qanet and \deca with pre-trained GloVe embeddings 
\cite{Pennington+:2014} and keep them fixed during training. We also 
freeze the input embeddings for \bert.\footnote{The input embeddings of 
\bert is a sum of token, segment and position embeddings; we freeze only 
the token embeddings.}  
To measure performance, we  use the standard 
macro-averaged F1 as the evaluation metric, which measures the average 
overlap of word tokens between prediction and ground truth 
answer.\footnote{If there are multiple ground truths, the maximum F1 is 
taken.} Our pre-trained \qanet, \deca and \bert achieve an F1 score of 
80.47, 75.50 and 87.62 respectively on the development partition of 
   \squad.  Note that the test partition of \squad is not released 
publicly, and so all reported \squad performance in the paper is on the 
development set.

Given the pre-trained \squad models, we fine-tune them on the \msmarco 
and \bioasq domains. We test vanilla fine-tuning (\finetune) and
 5 variants of fine-tuning with auxiliary penalty terms: (1) EWC 
(\ewc); normalised EWC (\ewcn); cosine distance (\cd); L2 (\ltwo); 
and combined normalised EWC, cosine distance and L2 (\all). As a 
benchmark, we also perform fine-tuning with gradient episodic memory 
(\gem), noting that this approach uses the first $m$ examples from 
\squad ($m = 256$ in our experiments).

 To find the best hyper-parameter configuration, we tune it based on the 
development partition for each target domain (and report the performance 
on the their test partitions).  
For a given domain, 
\finetune and its variants (\ewc, \ewcn, \cd, \ltwo and \all) all share 
the same hyper-parameter configuration.\footnote{The only exception are 
the scaling hyper-parameters ($\lambda_{ewc}$, $\lambda_{ewcn}$, 
$\lambda_{cd}$ and $\lambda_{l2}$), where we tune them separately for 
each model.} Detailed hyper-parameter settings are given in the 
supplementary material if requested.

As a baseline, we train \qanet, \deca and \bert from scratch (\scratch) 
using the target domain data. As before, we tune their hyper-parameters 
based on development performance. We present the full results in 
\tabref{rc-full}.

For each target domain, we display two F1 scores: the source \squad 
development performance (``\squad (dev)''); and the target domain's test 
performance (``Target (test)'')\footnote{Since we are evaluating different
auxiliary penalty terms in terms of their effectiveness in reducing catastrophic 
forgetting, we intentionally tuned the hyper-parameters for different auxiliary penalty terms
to generate similar performance on target domains. Therefore, we are more
interested in comparing the performance on the source domain, e.g. \squad (dev).  }. 
We first compare the performance between 
\scratch and \finetune. Across all domains for \qanet, \deca and \bert,   
\finetune substantially improves the target domain's performance 
compared to \scratch. The largest improvement is seen in \bioasq for 
\qanet, where its F1 improves two-fold (from 29.83 to 
65.81).  Among the three RC models, \bert has the
best performance for both \scratch and \finetune in most target domains 
(with a few exceptions such as \msfn and \mslw). Between \qanet and 
\deca, we see that \deca tends
 to have better \scratch performance but the pattern is reversed in 
 \finetune, where \qanet produces higher F1 than \deca in all domains 
except for \mslw.

In terms of \squad performance, we see that \finetune degrades it 
considerably compared to its pre-trained performance. The average
drop across all domains compared to their pre-trained performance is 
20.30, 15.30 and 15.07 points for \qanet, \deca and \bert, respectively.  
For most domains, F1 scores drop by 10-20 points, while for \mscp the 
performance is much worse for \qanet, with a drop of 41.34.
Interestingly, we see \bert suffers from catastrophic forgetting just as 
much as the other models, even though it is a larger model with orders 
of magnitude more parameters.

We now turn to the fine-tuning results with auxiliary penalties (\ewc, 
\ewcn, \cd and \ltwo).  Between \ewc and \ewcn, the normalised versions 
consistently produces better recovery for the source domain (one 
exception is \msms for \deca), demonstrating that normalisation helps.  
Between \ewcn, \cd and \ltwo, performance among the three models vary 
depending on the domain and there's no clear winner.  Combining all of 
these losses (\all) however, produces the best \squad performance for 
all models across most domains. The average recovery (\all - \finetune) 
of \squad performance is 4.54, 3.93 and 8.77 F1 points for \qanet, \deca 
and \bert respectively, implying that \bert benefits from these 
auxiliary penalties more than \deca and \qanet.

When compared to \gem, \all preserves \squad 
performance substantially better, on average 2.86 points more for \qanet 
and 5.57 points more \bert. For \deca, the improvement is minute (0.02).
Also, as \gem requires partial access to the training data from the source domain (\squad 
training examples in this case), the auxiliary penalty techniques are 
more favourable for real world applications. 

Does adding these penalty terms harm target performance?  Looking at the 
``Test'' performance between \finetune and \all, we see that they are 
generally comparable. We found that the average performance difference 
(\all -\finetune) is 0.23, $-$0.42 and 0.34 for \qanet, \deca and \bert 
respectively, implying that it does not (in fact, it has a small 
positive net impact for \qanet and \bert). In some cases it improves 
target performance substantially, e.g.\ in \bioasq for \bert, the target 
performance is improved from 71.62 to 76.93, when \all is applied.  

Based on these observations, we see benefits for incorporating these 
penalties when adapting comprehension models, as it produces a more 
robust model that preserves its source performance (to a certain extent) 
without trading off its target performance. In some cases, it can even 
improve the target performance.


\subsection{Continuous Learning}

In previous experiments, we fine-tune a pre-trained model to each domain 
independently. With continuous learning, we seek to investigate the 
performance of \finetune and its four variants (\ltwo, \cd, \ewcn and 
\all) when they are applied to a series of fine-tuning on multiple 
domains. For the remainder of this paper, we experiment only 
with \deca.

We have one model for each of the five fine-tuning methods, e.g. \finetune, 
\ltwo, \cd, \ewcn, and \all. 
Including the pre-training on \squad, all five models are trained for a total 
of 170K iterations: \squad (0--44K), \msbm (45K--65K), \mscp  
(66K--86K), \msfn (87K--107K), \msms (108K--128K), \msfm 
(129K--149K) and \mslw (150K--170K). 
When computing the penalties, we consider the trained model for the previous domain as the 
source model.\footnote{The implication is that  we have to re-compute 
the Fisher matrix for the last domain before we fine-tune the model on a 
new domain.} 
Figure \ref{fig:continuous} (a)-(c) demonstrate the performance of 
the five models on the development set of \squad and test sets of
 \msbm and \mscp, respectively, when they are adapted to 
 \msbm, \mscp, \msfn, \msms, \msfm and \mslw in sequence.\footnote{In 
 terms of hyper-parameters, we 
choose a configuration that is generally good for most domains.} We 
exclude plots for the latter domains as they are similar to that of \mscp. 

We first look at the recovery for \squad in \figref{cont_squad}.  \all 
(black line; legend in \figref{cont_comp}) trails well above all other 
models after a series of fine-tuning, followed by \ewcn and \cd, while 
\finetune produces the most forgetting. At the end of the continuous 
learning, \all recovers more than 5 F1 points compared to \finetune. We 
see a similar trend for \msbm (\figref{cont_bio}), although the 
difference is less pronounced.  The largest gap between \finetune and 
\all occurs when we fine-tune for \msfm (iteration 129K--149K).  Note 
that we are not trading off target performance when we first tune for 
\msbm (iteration 45K--65K), where \finetune and \all produces comparable 
F1.

For \mscp (\figref{cont_comp}), we first notice that there is 
considerably less forgetting overall (\mscp performance ranges from 
65--75 F1, while \squad performance in \figref{cont_squad} ranges from 
45--75 F1). This is perhaps unsurprising, as the model is already 
generally well-tuned (e.g.\ it takes less iterations to reach optimal 
performance for \mscp compared to \msbm and \squad). Most models perform 
similarly here. \all produces stronger recovery when fine-tuning on \msfm 
(129K--149K) and \mslw (150K--170K).  At the end of the continuous learning, 
the gap between all models is around 2 F1 points.

%

\subsection{Task Transfer}

\begin{table}
\begin{center}
\begin{adjustbox}{max width=\linewidth}
\begin{tabular}{ccccccccc}
\toprule
\textbf{Partition} & \textbf{Task} &
\textbf{\finetune} &\textbf{\ewc} & \textbf{\ewcn} 
&\textbf{\cd} & \textbf{\ltwo} &  \textbf{\all} \\
\midrule
\multirow{5}{*}{\squad (dev)}
& SUM & 8.60 & 11.65 & 12.48 & 11.28 & 9.34 & \bb{14.00}\\
& SRL & 50.51 & 51.30 & 56.99 & 55.40 & 51.56 & \bb{57.64}\\
& SP & 6.95  & 9.69 & 10.20 & 10.61 &19.39 &\bb{28.36} \\
& MT & 3.55 & 4.03 & \bb{4.29} & 3.48 & 3.15 & 3.59 \\
& SA & 1.74 & 2.69 & 2.38 & 3.63 & 2.51 & \bb{6.43}\\

\midrule

\multirow{5}{*}{Target (test)}
& SUM & 20.06 & 19.79 & 19.99 & 20.01 & \underline{20.38} & 20.12\\
& SRL & 71.69 & 71.80 & 71.74 & 72.12 & 71.90 & \underline{72.56}\\
& SP & 92.52 & \underline{92.77} & 92.70 & 92.62 & 92.59 & 91.11 \\
& MT & 24.99 & \underline{25.10} & 25.04 & 25.00 & 24.90 &24.90\\
& SA & 84.79 & \underline{86.38} & 84.84 & 85.06 &  86.27 & 85.89\\

\bottomrule
\end{tabular}
\end{adjustbox}
\end{center}
\caption{\deca's \squad and target performance for several tasks.}
\label{tab:deca-tasks}
\end{table}

In \deca, curriculum learning was used to train models for different NLP 
tasks. More specifically, \deca was first pre-trained on \squad and then 
fine-tuned on 10 tasks (including \squad) jointly.  During the training 
process, each minibatch consists of examples from a particular task, and 
they are sampled in an alternating fashion among different tasks.

In situations where we do not have access to training data from previous 
tasks, catastrophic forgetting occurs when we adapt the model for a new 
task. In this section, we test our methods for task transfer (as opposed 
to domain transfer in previous sections). To this end, we experiment 
with \deca and monitor its \squad performance when we fine-tune it for 
other tasks, including semantic role labelling (SRL), summarisation 
(SUM), semantic parsing (SP), machine translation (MT), and sentiment 
analysis (SA)\footnote{We did not test task transfer on \qanet or \bert
 because they are designed only for reading comprehension. }. 
Note that we are not doing joint or continuous learning 
here: we are taking the pre-trained model (on \squad) and adapting it to 
 new tasks independently. Description of these tasks are detailed in 
\cite{mccann+:2018}.

A core novelty of \deca is that its design allows it to generate words 
by extracting them from the question, context or its vocabulary, and 
this decision is made by the pointer-generator network. Based on the 
pointer-generator analysis in \cite{mccann+:2018}, we know that the 
pointer-generator network favours generating words using: (1) context 
for SRL, SUM, and SP; (2) question for SA; and (3) vocabulary for MT.

As before, \finetune serves as our baseline, and we have 5 variants with 
auxiliary penalty terms.  \tabref{deca-tasks} displays the F1 
performance on \squad and the target task; the table shares the same 
format as \tabref{rc-full}. 
In terms of target task performance (``Test''), we see similar 
performances for all five models. This is a similar observation we saw in 
previously, and it shows that the incorporation of the auxiliary penalty 
terms does not harm target task or domain performance.
For the source task \squad, \all produces substantial recovery for SUM, 
SRL, SP and SA, but not for MT. We hypothesise that this is due to the 
difference in nature between the target task and the source task:  i.e.\  
for SUM, SRL and SP, the output is generated by selecting words from 
context, which is similar to \squad; MT, on the other hand, generate 
using words from the vocabulary and question, and so it is likely to be 
difficult to find an optimal model that performs well for both 
tasks.

\section{Discussion}

\subsection{Gradient angle analysis}

The improved performance achieved by the combined auxiliary loss 
leads us to question that why it performed better than individual penalty
losses. An intuitive explanation is that different distance 
penalties, e.g. $l1$-distance, $l2$-distance and vector angles, form different 
constraints to prevent the model from being tuned too far away from the 
pre-trained model. Combining these distance penalties result in a more
strict constraints that helps further reducing the catastrophic forgetting. 
In addition to this, we hypothesise that the combined loss may be better at 
minimising the angle between the gradient vector w.r.t the target domain 
data and the gradient vector w.r.t the source domain data, during the 
process of fine-tuning, which is essentially the rationale behind the \gem
method \cite{lopez2017gradient}. 
 
\begin{figure}[t]
    \centering
        \includegraphics[width=0.9\linewidth]{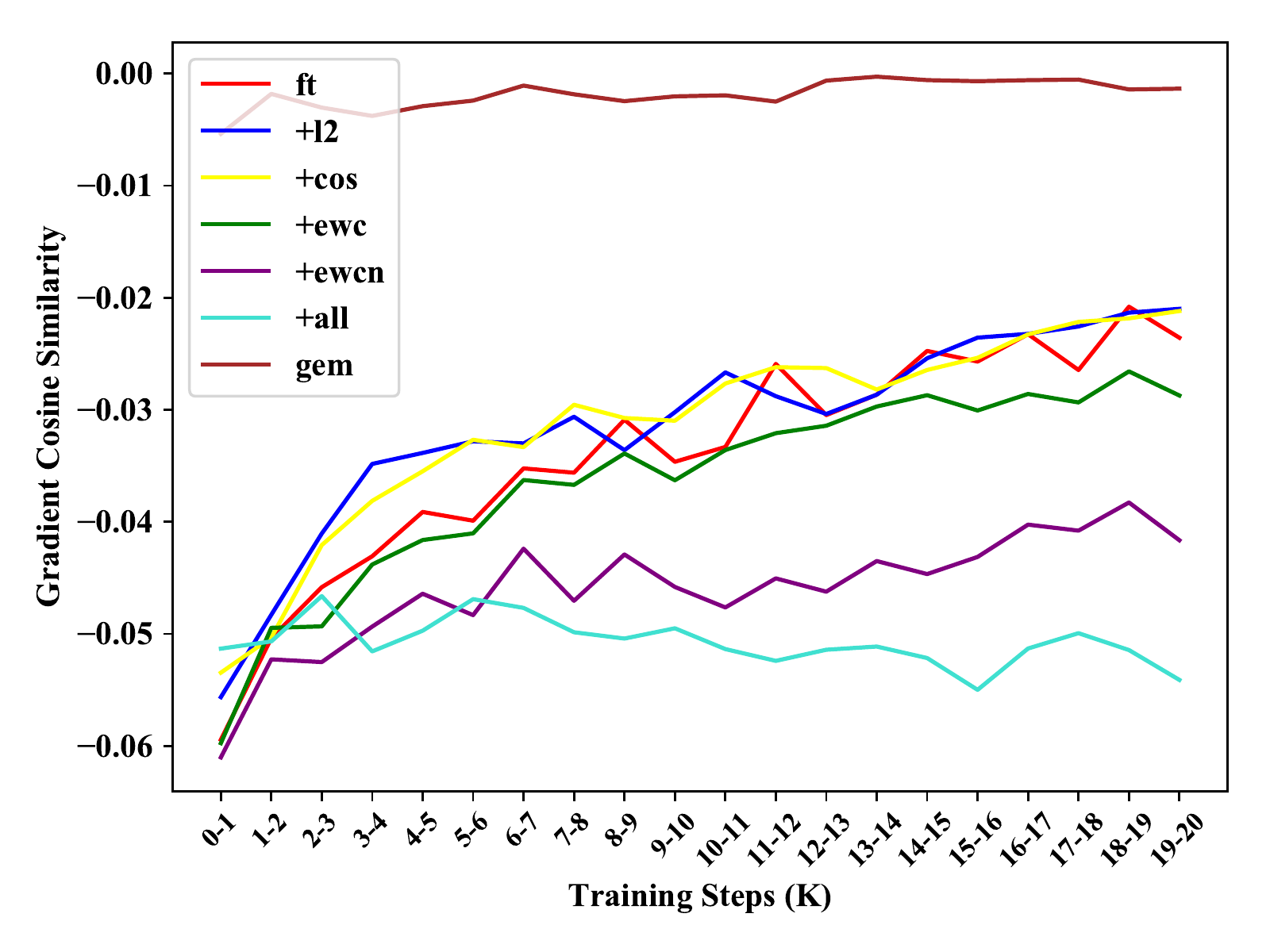}
        \caption{Averaged gradient cosine similarities on \msfn. }
        \label{fig:gradient-analysis}
\end{figure}

To validate this hypothesis, we conduct gradient analysis for all five
auxiliary penalty terms.  During fine-tuning, at each step $t$, we calculate 
the gradient cosine similarity $sim(g_t, g_t')$,   

\begin{align*}
    g_t&=\frac{\partial  \mathcal{L}(f_{\theta_t}, M)}{\partial \theta_t}
\\
    g_t'&=\frac{\partial  \mathcal{L}(f_{\theta_t}, (x, y))}{\partial \theta_t}
\end{align*}

\noindent where $M$ is a memory containing \squad examples, and $x$/$y$ is training 
data/label from the current domain.  
According to \cite{lopez2017gradient}, when the gradient cosine similarity is $\geq$0, 
the fine-tuning process is unlikely to harm the performance of the model on original 
domain. 
We smooth the scores
by averaging over every 1K steps, resulting in 20 cosine similarity 
values for 20K steps.  Figure \ref{fig:gradient-analysis} plots the 
gradient cosine similarity scores for the five models in \msfn. 

Curiously, our best performing model \all produces the lowest cosine 
similarity at most steps (the only exception is between 0-1K steps).  \finetune, 
on the other hand, maintains relatively high similarity throughout.  
Similar trends are found for other domains.
These observations imply that the approach \gem adopted --- i.e.\ 
constraining a positive dot product between $g_t$ and $g_t'$ --- is 
sufficient but not necessary for reducing catastrophic forgetting. 


\subsection{Decaying auxiliary loss scale}

Conventionally, the $\lambda$ scaling hyper-parameter for controlling 
the contribution of the penalty terms has a static value. In preliminary 
experiments, we notice that this loss starts at a very low value (close 
to zero), as initially there is little change to the model parameters.  
As such in the early iterations of fine-tuning, the model tends to focus 
on optimising for the target domain/task, and that results in a sharp 
drop of performance for the source domain/task.

In light of that, we explore using a dynamic $\lambda$ scale that starts 
at a larger value that decays over time.  With just simple linear decay,
 we found substantial improvement in \ewc for recovering \squad's 
performance, although the results are mixed for other penalties 
(particularly for \ewcn). We therefore only report results that are 
based on static $\lambda$ values in this paper. With that said, we 
contend that this might be an interesting avenue for further research, 
e.g.\ by exploring more complex decay functions.

\section{Conclusion}

To reduce catastrophic forgetting when adapting comprehension models, we 
explore several auxiliary penalty terms to regularise the fine-tuning process. 
We experiment with selective and non-selective penalties, and 
found that a combination of them consistently produces the best recovery 
for the source domain without harming its performance in the target 
domain. 
We also found similar observations when we apply our approach 
for adaptation to other tasks, demonstrating its general applicability.  
To test our approach, we develop and release six narrow domain reading 
comprehension data sets for the research community.

\bibliography{paper}
\bibliographystyle{paper}

\end{document}